\documentclass{article}
\usepackage{ijcai24}

\usepackage{times}
\usepackage{soul}
\usepackage{url}
\usepackage[hidelinks]{hyperref}
\usepackage[utf8]{inputenc}
\usepackage[small]{caption}
\usepackage{graphicx}
\usepackage{amsmath}
\usepackage{amsthm}
\usepackage{booktabs}
\usepackage{algorithm}
\usepackage{algorithmic}
\usepackage[switch]{lineno}

\usepackage{graphicx}
\usepackage{subfigure}
\usepackage{multirow}
\usepackage{makecell}

\title{MLLMReID: Multimodal Large Language Model-based Person Re-identification}

\author{
Shan Yang$^1$
\and
Yongfei Zhang$^2$\footnote{Corresponding Author}\\
\affiliations
$^1$Beijing Key Laboratory of Digital Media, School of Computer Science and Engineering,\\
Beihang University, Beijing 100191, China\\
$^2$State Key Laboratory of Virtual Reality Technology and Systems,\\ Beihang University, Beijing 100191, China\\
\emails
\{ shanyang, yfzhang\}@buaa.edu.cn
}

\begin{document}

\maketitle

\begin{abstract}
Multimodal large language models (MLLM) have achieved satisfactory results in many tasks. However, their performance in the task of ReID (ReID) has not been explored to date. This paper will investigate how to adapt them for the task of ReID. An intuitive idea is to fine-tune MLLM with ReID image-text datasets, and then use their visual encoder as a backbone for ReID. However, there still exist two apparent issues: (1) Designing instructions for ReID, MLLMs may overfit specific instructions, and designing a variety of instructions will lead to higher costs. (2) When fine-tuning the visual encoder of a MLLM, it is not trained synchronously with the ReID task. As a result, the effectiveness of the visual encoder fine-tuning cannot be directly reflected in the performance of the ReID task. To address these problems, this paper proposes MLLMReID: Multimodal Large Language Model-based ReID. Firstly, we proposed Common Instruction, a simple approach that leverages the essence ability of LLMs to continue writing, avoiding complex and diverse instruction design. Secondly, we propose a multi-task learning-based synchronization module to ensure that the visual encoder of the MLLM is trained synchronously with the ReID task. The experimental results demonstrate the superiority of our method.
\end{abstract}

\section{Introduction}
\begin{figure*}[h]
\centering
\subfigure[Multimodal Large Language Model Structure Diagram.]{
\includegraphics[scale=0.3]{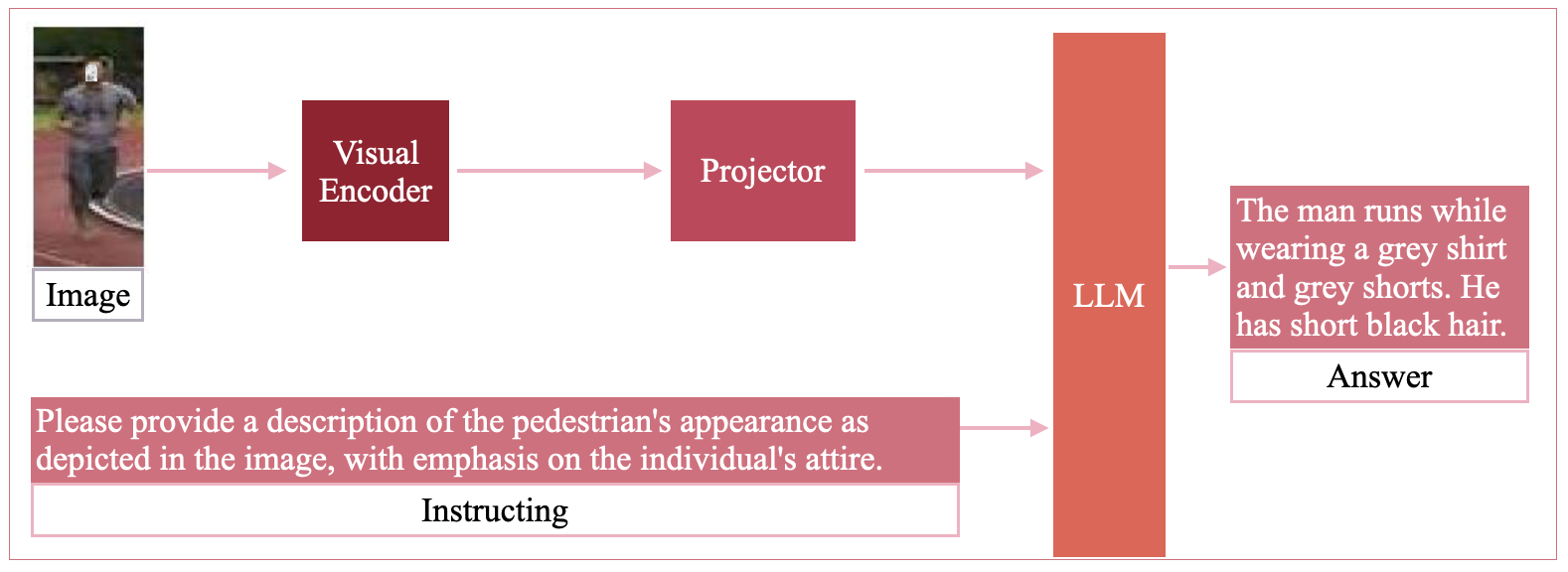}
\label{llava}
}
\subfigure[Person Re-identification Model Structure with Fine-Tuned Visual Encoder.]{
\includegraphics[scale=0.24]{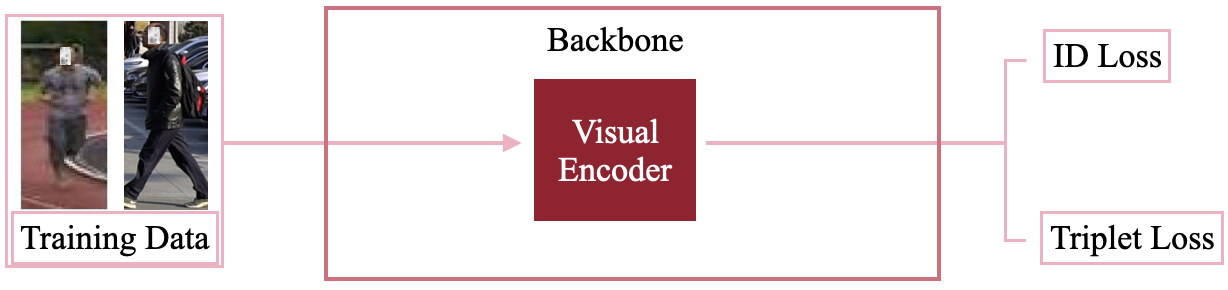}
\label{llava-reid}
}
\caption{Naive Implementation of Multimodal Large Language Model for Person Re-identification.}
\label{baseline}
\end{figure*}

With the advancement of Large Language Models (LLM)~\cite{touvron2023llama}, Multimodal Large Language Models(MLLM) have achieved excellent results in various tasks~\cite{Yin2023ASO,Li2023MultimodalFM}, such as multimodal question answering~\cite{Liu2023VisualIT} and embodied agents~\cite{Brohan2023RT2VM}. However, the effectiveness of MLLM in ReID (ReID) tasks remains unexplored. ReID refers to the cross-camera association of target persons~\cite{Ye2020DeepLF}. This paper focuses on how to make MLLM effective in ReID tasks. An intuitive idea is to fine-tune a MLLM, such as LLaVA~\cite{Liu2023VisualIT}, using ReID image-text pairs, as shown in Figure~\ref{llava}. During fine-tuning, a projection layer is used for multimodal alignment of image and text, while indirectly optimizing the visual encoder through the LLM's excellent representational capabilities, thereby enhancing feature extraction for persons and improving task performance. The fine-tuned visual encoder is then used as the backbone network for the ReID model, followed by training the ReID model, illustrated in Figure~\ref{llava-reid}. Although this idea may be simple and effective, how to better adapt MLLMs to ReID tasks still poses two urgent challenges.

Firstly, when applying MLLMs to ReID tasks, designing instructions is inevitable. Using instructive learning results in the model adapting to the given instruction, leading to overfitting. Lora~\cite{Hu2021LoRALA} fine-tuning may cause the LLM to lose some diversity, leading to poorer generalization on unseen samples due to overfitting to the instruction. To preserve diversity and enhance generalizability, generating more comprehensive instructions is intuitive, but this poses several issues: (1) It's difficult to generate high-quality and diverse instructions; (2) Even with an abundance of instructions, appropriate training data matching these instructions is required; (3) The richness of instructions leads to a geometric increase in training costs. Therefore, these issues result in additional costs, making it our goal to find a more common instruction that doesn't change the nature of the LLM. The essence of an LLM is to naturally generate subsequent sentences based on previous tokens. We design a simple and common instruction to enable image and text to produce the same continuation text, thereby solving this problem.

Secondly, the visual encoder of the MLLM and the ReID task are not trained synchronously. Specifically, the latent image feature vectors output by the LLM are not used to compute loss in instruct learning~\cite{Liu2023VisualIT,Zhu2023MiniGPT4EV}, leading to indirect optimization of multimodal features and insufficient utilization of these features. This approach is not efficient and is not conducive to learning by the visual encoder, thus impairing feature extraction for persons. Additionally, when fine-tuning the visual encoder of the MLLM, it is isolated from the ReID task, meaning that the effectiveness of the visual encoder fine-tuning cannot be directly reflected in the ReID task. Therefore, we directly apply the latent image feature vectors outputted by LLMs to the task of ReID, utilizing the loss generated in this task to directly optimize the visual encoder or projection layer.

To address the aforementioned challenges, this paper proposes MLLMReID: Multimodal Large Language Model-based ReID. Firstly, it introduces the concept of Common Instruction for leveraging the essence of LLM in continuation writing. This is a simple continuation instruction designed to avoid the high costs arising from complex and diverse instruction designs. Secondly, the Multi-Task Learning-based Synchronization (SyncReID) module proposed in this paper enables the visual encoder of the MLLM to be trained synchronously with the ReID task.

Here's a summary of our contributions:

(1) We created common instructions to bypass the need for complex, image-caption-specific ones, improving the generalizability of our method on ReID datasets with image-caption data.

(2) We optimized the visual encoder using latent image features from the Large Language Model (LLM), training the visual encoder synchronously with the ReID task. This greatly enhances the MLLM's ability to understand and extract person features.

(3) We conducted experiments to demonstrate the effectiveness of applying MLLM to ReID tasks.

\begin{figure*}[h]
    \centering
    \includegraphics[scale=0.27]{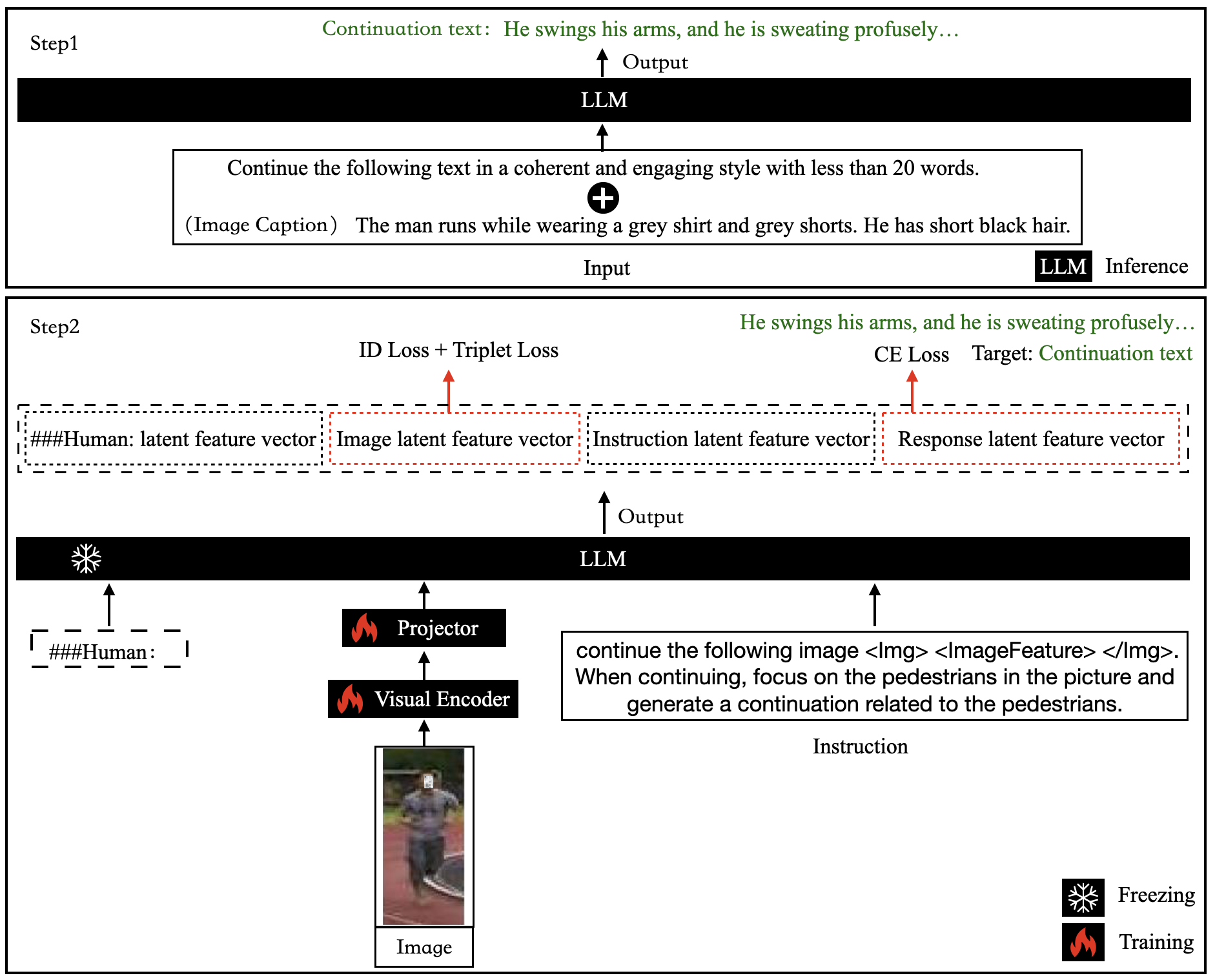}
    \caption{Overview of our proposed approach: Multimodal Large Language Model-based Person Re-identification.}
    \label{mllmreid}
\end{figure*}
\section{Related Work}
\subsection{Image-Text Multimodal person re-identification}
Existing methods use textual commands or multimodal text-image input to guide the model in learning person features within images. UniHCP~\cite{Ci2023UniHCPAU} utilizes a query to focus the model on cloth-agnostic features, such as "extract cloth-agnostic identity representations, dim: 1536," within an encoder-decoder Transformer architecture. This model emphasizes features that are independent of clothing. In common ReID tasks, the focus should ideally be on persons' clothing to distinguish different people. If the model centers on cloth-agnostic features, it might struggle to learn discriminative attributes. Moreover, though the query contains explicit command instructions, there is no corresponding supervisory information within the framework, possibly rendering the query ineffective. In contrast, our method includes specific text descriptions corresponding to command instructions as supervisory information, ensuring the practical impact on the model. \cite{Bao2023LearningTP} introduced VAL-PAT, utilizing self-supervised contrastive learning, image-text contrastive learning, and multi-attribute classification for unsupervised ReID. However, using attribute-based training can pose two challenges: (1) Some fine-grained attribute information, such as glasses, is hard to manifest in images down-sampled to sizes like 256*128 or 384×128, typically used to increase retrieval efficiency, and the model may find it challenging to extract this information. (2) Utilizing attributes in feature learning is complex and inflexible, as it necessitates memorizing numerous labels and may fail to capture diverse variations in human appearance adequately~\cite{Chen2018ImprovingDV}. Some methods employ CLIP's visual encoder and text encoder~\cite{He2023RegionGA,Chen2023TowardsMP}, yet the contrastive learning approach of CLIP can lead to poor quality of visual encoder representations due to misalignment between image and text concepts, ultimately affecting the performance of ReID models. In contrast, we are the first to explore the application of MLLM in ReID. We propose common instructions and demonstrate how to effectively utilize the latent features of images outputted by LLM to enhance the performance in ReID tasks.

\section{Method}
We introduce Common Instruction fully leverages the continuation characteristics of the LLM, allowing both text and image inputs to generate identical continuation texts. It simplifies the process by requiring only straightforward continuation instructions, thereby avoiding the challenges associated with designing diverse commands. To enhance the performance of the visual encoder in ReID tasks, the proposed SyncReID module utilizes the latent image feature vectors outputted by the LLM for ReID tasks, subsequently optimizing the visual encoder.

\subsection{Architecture of MLLM}
To achieve our main objective, we skillfully utilize the strengths of the pre-existing LLM and the visual model, depicted in Figure~\ref{mllmreid}. We have selected LLaMA 2 for our LLM, owing to its proven efficiency in various publicly available projects focusing solely on language-based instruction tuning~\cite{touvron2023llama}. For the visual input, we utilize the pre-trained CLIP visual encoder, specifically the ViT-L/14 variant~\cite{Dosovitskiy2020AnII}, to derive the visual representation $f_v$. Our research includes examining the feature maps at stages both preceding and succeeding the final Transformer layer. To bridge the gap between image features and the linguistic embedding space, we employ a straightforward linear layer. More precisely, we use a modifiable projection matrix to transform $f_v$ into language embedding tokens. These tokens are configured to match the dimensionality of the word embedding space found in the language model.
\subsection{Common Instruction}
Common Instruction leverages the essence of LLM continuation to design a common instruction that enables both text and image inputs to produce identical continuation texts.

Firstly, continuation texts are generated using the caption text from image-text pair data, with the following instruction for continuation:\\
\textit{Continue the following text in a coherent and engaging style with less than 20 words.}\\
This process only requires inputting the caption text into the LLM, which, upon receiving the continuation instruction, generates the corresponding continuation text. It is important to note that this step involves only inference.

Secondly, the image from the image-text pair data is used to generate the continuation text, with the instruction as follows:\\
\textit{\#\#\#Human: continue the following image $<$Img$>$ $<$ImageFeature$>$ $<$/Img$>$. When continuing, focus on the persons in the picture and 
generate a continuation related to the persons.\\
\#\#\#Assistant:\\}
Here, the image is input into the visual encoder, and upon receiving the continuation instruction, the LLM outputs the corresponding continuation text. Notably, this step involves fine-tuning the MLLM, where the target is the continuation text rather than the caption.
For every image Xv, we create a series of multi-turn dialogue data $\left(\mathbf{X}_{\mathrm{q}}^1, \mathbf{X}_{\mathrm{a}}^1, \cdots, \mathbf{X}_{\mathrm{q}}^T, \mathbf{X}_{\mathrm{a}}^T\right)$, with T representing the complete count of dialogue exchanges.

We conduct instruction-based tuning on the LLM for the prediction tokens, adhering to its inherent auto-regressive training goal. In particular, for a sequence with length $L$, we calculate the likelihood of producing the desired answers $Xc$ by:
\begin{equation}
p\left(\mathbf{X}_{\mathrm{c}} \mid \mathbf{X}_{\mathrm{v}}, \mathbf{X}_{\text {instruct }}\right)=\prod_{i=1}^L p_{\boldsymbol{\theta}}\left(x_i \mid \mathbf{X}_{\mathrm{v}}, \mathbf{X}_{\text {instruct },<i}, \mathbf{X}_{\mathbf{c},<i}\right),
\end{equation} 
where $\theta$ represents the parameters that can be trained. $\mathbf{X}_{\text {instruct },<i}$ and $\mathbf{X}_{\mathbf{c},<i}$ denote the tokens for instruction and continuation text in all preceding turns up to the current prediction token $x_i$, respectively. 

This process effectively employs the Knowledge Distillation strategy~\cite{Hinton2015DistillingTK}. It ensures that both text inputs (captions) and image inputs produce identical continuation texts. In other words, regardless of the type of input, the output of the LLM remains consistent. Furthermore, the continuation text is an instruction that aligns perfectly with the inherent nature of the LLM. There is no need to design complex and diverse instructions; rather, only two simple, universal continuation instructions are required to fine-tune the MLLM.

\subsection{SyncReID}
In this paper, the latent feature vectors of images outputted by the LLM are applied to the task of ReID. This not only effectively utilizes these features but also directly optimizes the visual encoder or projection layer, which is beneficial for the extraction of person features.

To achieve this, we employ the latent feature vectors of images in the classical loss functions used in ReID: ID Loss and Triplet Loss.

\begin{equation}
L_{\mathrm{ID}}(\theta ; x)=-\frac{1}{P \times K} \sum_{i=1}^{P \times K} \log \left(p\left(y_i \mid x_i\right)\right)
\end{equation}

Where $P*K$ is the batch size, $x_i$ refers to the image sample of person $i$, and $y_i$ denotes its category label.

\begin{equation}
\begin{aligned}
L_{\mathrm{tri}}(\theta ; X) & =\sum_{i=1}^P \sum_{a=1}^K\left[m+\max _{p=1 \ldots K} D\left(f_\theta\left(x_a^i\right), f_\theta\left(x_p^i\right)\right)\right. \\
& \left.-\min _{j=1 \ldots P, n=1 \ldots K, j \neq i} D\left(f_\theta\left(x_a^i\right), f_\theta\left(x_n^j\right)\right)\right]_{+}
\end{aligned}
\end{equation}

Where $f_\theta\left(x_a^i\right)$ refers to the anchor sample, $f_\theta\left(x_p^i\right)$ indicates the hard positive sample and $f_\theta\left(x_n^i\right)$ denotes the hard negative sample.

\subsection{Overall Loss}
To balance two types of loss functions—the cross-entropy (CE) loss generated by the LLM in predicting the next token, and the loss associated with the ReID task—we introduce a parameter $\lambda$ for balance. 
\begin{equation}
\begin{aligned}
L_{\text {over }}(\theta ; x) & =\lambda \times p\left(\mathbf{X}_{\mathrm{c}} \mid \mathbf{X}_{\mathrm{v}}, \mathbf{X}_{\text {instruct }}\right)+ \\
& (1-\lambda) \times\left(L_{\mathrm{ID}}(\theta ; x)+L_{\text {tri }}(\theta ; X)\right)
\end{aligned}
\label{lambda_equ}
\end{equation}

\section{Experiments}
\begin{table*}
\begin{center}
\setlength{\tabcolsep}{7mm}{
\begin{tabular}{llllll}
\hline
\multirow{2}{*}{Backbone} & \multirow{2}{*}{Method}  &  \multicolumn{2}{c}{MSMT17}  & \multicolumn{2}{c}{Market1501}  \\
\cline{3-6}
 & & mAP & R1 & mAP & R1\\
\hline
\multirow{8}{*}{CNN} 
& CBN~\cite{zhuang2020rethinking} & 42.9 & 72.8 & 77.3 & 91.3 \\

 & MGN~\cite{wang2018learning}& 52.1 & 76.9 & 86.9 & 95.7 \\
 
 & SCSN~\cite{chen2020salience}  & 58.5 & 83.8 & 88.5 & 95.7  \\
 & ABDNet~\cite{chen2019abd}  & 60.8 & 82.3 & 88.3 & 95.6 \\
 & PGFA~\cite{miao2019pose} & - &  - &  76.8 &  91.2  \\
 & HOReID~\cite{wang2020high} & - & - & 84.9 & 94.2 \\
 & ISP~\cite{zhu2020identity} & - & - & 88.6 & 95.3  \\

\hline
\multirow{3}{*}{ViT-B/16} 
 & TransReID~\cite{he2021transreid} & 69.4 & 86.2 & 89.5 & 95.2 \\
 & UniHCP~\cite{Ci2023UniHCPAU} & 67.3 & - & 90.3 & - \\
 & RGANet~\cite{He2023RegionGA} & 72.3 & 88.1 & 89.8 & 95.5 \\
  & CLIP-ReID~\cite{{li2023clip}} &  74.3 & \textbf{88.9} & 90.7 & 95.8\\
\hline

\multirow{2}{*}{ViT-L/14}

 & Baseline & 72.7 & 87.0 & 90.1 & 95.6  \\
 & MLLMReID (Ours) & \textbf{76.7} & 87.9 & \textbf{91.5} & \textbf{96.5}  \\
\hline
\end{tabular}}
\end{center}

{$^{\mathrm{*}}$Note: In the testing phase, both the Baseline and our method MLLMReID solely utilize the visual encoder, without employing LLM or textual descriptions. The experimental results of CLIP-ReID in the table were reproduced using the official code provided by CLIP-ReID~\cite{{li2023clip}}.}
\caption{Comparing the Performance of Different Methods Evaluated on Public Datasets$^{\mathrm{*}}$}
\label{reidexp}
\end{table*}

\subsection{Experiment Setting Overview}
\textbf{Datasets: }In multimodal pretraining, we employed annotations from the CUHK-PEDES~\cite{Li2017PersonSW} dataset to augment the Market1501~\cite{zheng2015scalable} and CUHK03~\cite{li2014deepreid} datasets. As the RSTPReid~\cite{Zhu2021DSSLDS} and ICFG-PEDES~\cite{Ding2021SemanticallySN} datasets are derived from MSMT17~\cite{wei2018person} for image-text annotation, we aligned the textual annotations of images in these two datasets with the original MSMT17 dataset. For images lacking text descriptions, we followed the protocol in~\cite{Chen2018ImprovingDV}, extending descriptions from images of the same person under the same camera to those images as much as possible. Building upon the aforementioned datasets, we will now delve into the specific details of each, including the division of content and the unique characteristics that distinguish them. The CUHK-PEDES~\cite{Li2017PersonSW} dataset encompasses 40,206 images representing 13,003 persons, along with 80,440 descriptive sentences. This dataset is divided into three subsets: the training subset contains 34,054 images and 68,126 descriptive sentences; the validation subset consists of 3,078 images and 6,158 descriptive sentences; the testing subset holds 3,074 images and 6,156 descriptive sentences. As for RSTPReid~\cite{Zhu2021DSSLDS}, gathered from MSMT17~\cite{wei2018person}, it includes 20,505 images of 4,101 persons. With every image, there are two sentences, each having no fewer than 23 words. In further detail, the training, validation, and testing subsets contain 3,701, 200, and 200 identities, correspondingly. Additionally, ICFG-PEDES~\cite{Ding2021SemanticallySN}, also derived from MSMT17~\cite{wei2018person}, incorporates 54,522 images of 4,102 persons. Accompanying each image is a descriptive sentence averaging 37.2 words. The training and testing portions encompass 34,674 image-text pairs for 3,102 persons, and 19,848 image-text pairs for the remaining 1,000 persons, respectively. 

In the tasks of ReID and generalized ReID, we utilize four datasets to validate our proposed scheme: Market1501~\cite{zheng2015scalable}, DukeMTMC-ReID~\cite{ristani2016performance}, MSMT17~\cite{wei2018person}, and CUHK~\cite{li2014deepreid}. Table~\ref{tablea0} presents an all-encompassing statistical analysis of the multiple datasets employed in our research for ReID purposes. This summary includes essential details like the number of person IDs and images across the query, gallery, and training collections, along with the total camera count for each dataset. Insight into these datasets is vital for comprehending the structure of our experiments and provides a foundation for subsequent interpretation of our findings.
\begin{table}

\begin{center}
\setlength{\tabcolsep}{0.25mm}{
 \begin{tabular}{llllllll}        
            \hline
            Dataset & $ID_q$ & $ID_g$ & $ID_t$ & $IMG_q$ & $IMG_g$ & $IMG_t$ & $CAM_n$\\
            \hline
            MSMT17 & 3060 & 3060  & 1041 & 11659 & 82161 & 32621 & 15 \\
            Market1501  & 750 & 750  & 751 & 3368 & 19732 & 12936 & 6 \\
            DukeMTMC & 702  & 702   & 702  & 2228 & 17661 & 16522 & 8 \\
            CUHK03-NP  & 700 & 700  & 767 & 1400 & 5332 & 7365 & 2\\
            \hline
\end{tabular}}
\end{center}
{$^{\mathrm{*}}$$ID_q$, $ID_g$, and $ID_t$ represent the number of IDs in the query set, gallery set, and training set, respectively. $IMG_q$, $IMG_g$, and $IMG_t$ denote the number of images in the query set, gallery set, and training set. $CAM_n$ indicates the number of cameras in the dataset.}
\caption{Statistics of ReID Datasets.$^{\mathrm{*}}$}
\label{tablea0}
\end{table}

\textbf{Evaluation:} We employ Rank-1 (R1) and mean Average Precision (mAP) as our evaluation metrics. R1 measures the accuracy of the first result in the retrieval process, whereas mAP assesses the overall ranking accuracy of the retrieved results. Cross-dataset evaluation is conducted by training the model on the training set of the source dataset and then assessing it on the test set of the target dataset.

\textbf{Baseline System:} We adopt the model depicted in Figure~\ref{baseline} as our baseline system. Initially, we fine-tune LLaVA using image-text pair data, and then employ the fine-tuned visual encoder as the backbone for the ReID task.

To fulfill the design requirements for diverse instructional prompts, we have developed 20 different prompts that primarily focus on various inquiries related to the appearance and combinations of clothing. A selection of these prompts is exhibited in Table~\ref{instruct}. For our experiments, a prompt is randomly selected for each image-text pair. To ensure the reliability of the Baseline results reported in the following sections, we conducted ten experimental runs and calculated the average to reduce the impact of randomness.

\begin{table*}

    \centering
    \setlength{\tabcolsep}{0.5mm}{
    \begin{tabular}{cc}
    \hline
          Prompt & Example\\
    \hline
         Prompt1 & Describe the appearance of persons in the image, focusing on their attire. \\
        Prompt2 & Elucidate the visual features of persons in the image, including their dress style.\\
        Prompt3 &  \makecell{Provide a detailed interpretation of the persons' physical attributes in the image,\\ especially their clothing combinations.}\\
        Prompt4 & \makecell{Analyze the observable characteristics of persons in the image, \\specifically related to their clothing arrangements.}\\
        Prompt5 & Depict the physical features of persons in the image, such as their clothing combinations.\\
        Prompt6 & Analyze the appearance of persons in the image, with a focus on their clothing coordination.\\
    \hline          
    \end{tabular}}
    \caption{Examples of Prompts}
    \label{instruct}
\end{table*}

\textbf{Experiment Settings:} Our method is based on the official PyTorch code of TransReID\footnote{https://github.com/damo-cv/TransReID}~\cite{he2021transreid}, LLaVa\footnote{https://github.com/haotian-liu/LLaVA}, and GS\footnote{https://github.com/ShengcaiLiao/QAConv}~\cite{liao2020interpretable}. In the ReID task, we employ ViT-L/14~\cite{DBLP:conf/iclr/DosovitskiyB0WZ21} as our visual encoder, and LLaMA 2-7B~\cite{touvron2023llama} for LLM. Data augmentation strategies include random horizontal flipping, padding, random cropping, and random erasing. The backbone network for the generalized re-identification task is ViT-L/14~\cite{DBLP:conf/iclr/DosovitskiyB0WZ21}. Data augmentation strategies consist of random cropping, flipping, and occlusion. Other hyperparameters, such as batch size and learning rate, are set according to the configuration files of LLaVA, TransReID~\cite{he2021transreid}, and GS~\cite{liao2020interpretable}. Some table columns are left blank, indicating that the method has not been tested on the corresponding dataset yet. The parameter value $\lambda=0.3$, mentioned in the method section, is determined empirically. Our experiments were carried out on the NVIDIA A100 GPU using the PyTorch toolbox\footnote{http://pytorch.org}.

\textbf{Training Protocol:} When training the ReID model on the MSMT17 dataset, the pre-trained visual encoder utilize only the corresponding MSMT17 dataset that comes with text descriptions, adhering to this protocol even when training the ReID model on other datasets.

\subsection{Performance Comparison}

\subsection{Analyzing and Comparing Performance Metrics}
In Table~\ref{reidexp}, we present a comparison of various methods across three columns. The first column encompasses representative ReID techniques that primarily employ CNN-based backbone networks, while the second column features representative methods that use ViT as the backbone network, such as the recent state-of-the-art UniHCP~\cite{Ci2023UniHCPAU}, as well as RGANet~\cite{He2023RegionGA}, which employs CLIP image and text encoders. The baseline and our proposed method are showcased in the final column. 

The experimental results show that the proposed method achieves certain performance improvements compared to other listed methods. Notably, on the large-scale MSMT17 dataset, our method demonstrates significant progress, thereby validating its effectiveness. Compared to RGANet, our method improves the mAP score on MSMT17 by 4.4\%. Compared to CLIP-ReID, our method improves the mAP score on MSMT17 by 2.4\%. This improvement can be attributed to the Common Instruction ensuring that text and image inputs maintain the same output. This distillation design mechanism allows the visual encoder to more profoundly understand the appearance of persons in the images, thus aiding in the learning of fine-grained features. Furthermore, the addition of SyncReID, which effectively utilizes the LLM output's latent image features for the ReID task, further promotes our model's ability to learn distinctive fine-grained features. Compared to the baseline, our method improves the mAP and Rank-1 indicators on the MSMT17 dataset by 4.0\% and 0.9\%, respectively. This indicates that the proposed Common Instruction effectively leverages the essence of LLM continuation. Since it is merely a continuation instruction, it does not cause LLM overfitting on a specific instruction like designing diverse instructions, which would reduce the model's generalizability and be detrimental to MLLM learning, ultimately leading to a decline in the representational ability of the visual encoder. This shows that our Common Instruction has stronger generalizability. Meanwhile, our proposed SyncReID module can effectively utilize the latent image feature vectors output by LLMs, applying them to the ReID task. SyncReID optimizes the visual encoder using these features in ID Loss and Triplet Loss, thereby enhancing the encoder's capability to extract person features.

This achievement not only reveals the satisfactory performance of our method in the experiments but also further strengthens our anticipation and confidence in its potential applications.

\begin{table*}

\begin{center}
\setlength{\tabcolsep}{3.5mm}{
\begin{tabular}{llllllll}
\hline
\multirow{2}{*}{Method} & \multirow{2}{*}{Training set} & \multicolumn{2}{c}{CUHK03-NP} & \multicolumn{2}{c}{Market1501} & \multicolumn{2}{c}{MSMT17} \\
\cline{3-8}
 &  & R1 & mAP & R1 & mAP & R1 & mAP \\
\hline
M$^{3}$L~\cite{zhao2021learning} & Multi & 33.1 & 32.1 & 75.9 & 50.2 & 36.9 & 14.7 \\
\hline
MGN~\cite{wang2018learning} & Market1501 & 8.5 & 7.4 & - & - & - & - \\
CBN~\cite{zhuang2020rethinking}  & Market1501 & - & - & - & - & 25.3 & 9.5 \\
QAConv-GS~\cite{Liao_2022_CVPR} & Market1501 & 19.1 & 18.1 & - & - & 45.9 & 17.2 \\

TransMatcher~\cite{Liao2021TransMatcherDI} & Market1501  & 22.2 & 21.4 & - & - & 47.3 & 18.4 \\
MMET$^{\mathrm{*}}$~\cite{Xiang2023LearningRV} & Market1501 & 22.3 & 21.4 & - & - & 47.4 & 18.4 \\
MLLMReID (Ours) & Market1501 & \textbf{24.5} & \textbf{23.4} & - & - & \textbf{49.1} & \textbf{20.5} \\

\hline
PCB~\cite{sun2018beyond} & MSMT17 & - & - & 52.7 & 26.7 & - & - \\
QAConv-GS~\cite{Liao_2022_CVPR}  & MSMT17 & 20.9 & 20.6 & 79.1 & 49.5 & - & - \\
TransMatcher~\cite{Liao2021TransMatcherDI} & MSMT17 & 23.7 & 22.5 & 80.1& 52.0 & - & - \\
MMET$^{\mathrm{*}}$~\cite{Xiang2023LearningRV}& MSMT17  & 23.8 & 22.5 & 80.0 & 52.1 & - & - \\
MLLMReID (Ours) & MSMT17  & \textbf{25.4} & \textbf{24.0} & \textbf{81.4} & \textbf{53.1} & - & - \\


\hline
\end{tabular}}
\end{center}
{$^{\mathrm{*}}$Note: In the testing phase, both the Baseline and our method MLLMReID solely utilize the visual encoder, without employing LLM or textual descriptions. In our study, MMET$^{\mathrm{*}}$ signifies the replication of the existing method using the official codebase. By employing the official MMET code\footnote{https://github.com/JeremyXSC/MMET}, we successfully implemented the multi-modal ReID approach across a range of datasets, such as Market1501, CUHK03, and MSMT17. Our implementation rigorously follows the methodologies and configurations outlined in the papers~\cite{Shu2022SeeFS} and~\cite{Qi2023AnID}, ensuring fidelity to the original research while exploring its application in diverse dataset environments.}
\caption{Direct cross-dataset evaluation results.$^{\mathrm{*}}$}
\label{reidgen}
\end{table*}

\subsection{Performance Comparison in Generalizable ReID Task}
We choose to test our model on the generalizable ReID task to validate its robustness and adaptability when facing different environments and unseen data, which are crucial for practical applications. When comparing our proposed method with classical generalized methods listed in Table~\ref{reidgen}, it can be observed that our model outperforms these methods. The possible reasons are as follows:

First, the effective integration of the Common Instruction and the SyncReID module within our model architecture significantly contributes to its enhanced performance. The Common Instruction, by streamlining the process of handling both text and image inputs, ensures a consistent treatment of multimodal data, thereby improving the model's ability to generalize across different datasets. This is particularly important in scenarios where the model encounters diverse and previously unseen data, as is often the case in generalizable ReID tasks.

Second, the SyncReID module plays a pivotal role in harnessing the potential of LLMs for visual feature extraction. By directly applying the latent image feature vectors output by LLMs to the ReID task, the model is able to capture more nuanced and distinctive features of individuals, which is crucial for accurate ReID. This approach contrasts with traditional methods that may not fully exploit the rich feature representations provided by LLMs, thus limiting their effectiveness in varied and complex real-world environments.

In summary, our model's superior performance in the generalizable ReID task is a testament to its robust design, effective utilization of advanced LLM features, and comprehensive training. These factors collectively contribute to its strong adaptability and reliability in different environments, making it a promising solution for real-world ReID challenges.

\subsection{Ablation Study}

\begin{table}
\setlength{\tabcolsep}{3mm}{
\begin{tabular}{ccc}
\hline
 & Rank-1 (\%) & mAP (\%) \\
\hline
Baseline Model & 72.7  & 87.0 \\
+ Common Instruction & 74.6 & 87.3 \\
+  SyncReID & 75.3 & 87.2 \\
\makecell{+ Common Instruction \\+ SyncReID} & 76.7 & 87.9 \\
\hline
\end{tabular}}
\caption{Results of the ablation study on the MSMT17 dataset.}
\label{ablation_study}
\end{table}

In our ablation study on the MSMT17 dataset, we evaluated the impact of Common Instruction and SyncReID modules within the MLLMReID framework. The baseline model's performance was first established, which was then sequentially enhanced by integrating these modules. The addition of Common Instruction confirmed its efficacy in improving model performance. Integrating SyncReID led to further improvements, with notable increases in mAP and Rank-1 scores. The synergistic incorporation of both modules substantially elevated performance over the baseline, affirming the strength of our method in advancing MLLM application in ReID tasks.

\subsection{Case Study}
\begin{figure}[h]
    \centering
    \includegraphics[scale=0.175]{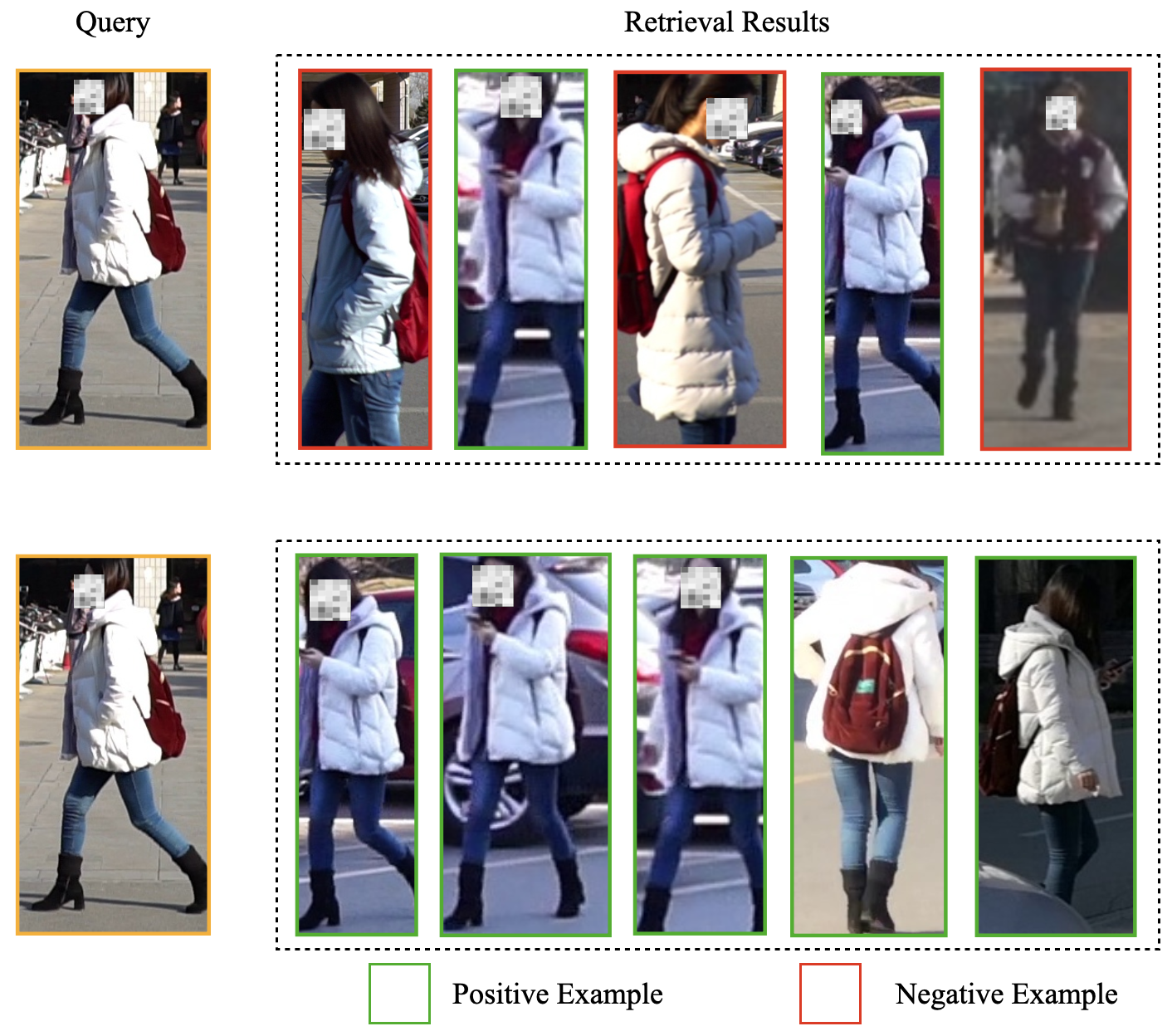}
    \caption{Schematic representation of ReID retrieval results. The first and second rows represent the retrieval results of the baseline system and our method, respectively.}
    \label{unit4_2retrival}
\end{figure}
In this section, we investigate the efficacy of our model through a specific case study. We evaluate the contributions of the Common Instruction and the SyncReID. Figure \ref{unit4_2retrival} contrasts the retrieval results of our approach with those of the baseline system. It is evident from the results that our method can effectively differentiate between persons with similar appearances. This distinction can be attributed to the following reasons:

Firstly, the Common Instruction's ability to harmonize the output from both text and image inputs. Through this distilled design approach, the visual encoder gains a deeper comprehension of the persons' appearance within the images, which facilitates the acquisition of detailed and nuanced features. By encouraging the model to concentrate on the granular aspects highlighted in the textual descriptions, the Common Instruction promotes a more focused and discerning feature analysis.

Secondly, the SyncReID module capitalizes on the direct application of latent image features from the LLM, which enriches the feature set that the model can utilize. By directly optimizing these features through the ReID task-specific loss functions such as ID Loss and Triplet Loss, the visual encoder is able to extract and learn a richer representation of person features.

The combination of these two strategies results in a more powerful and discriminative model. The retrieval results, as shown in Figure 3, demonstrate that our approach significantly reduces the chances of mismatching while increasing the retrieval accuracy. For instance, in the case study, the baseline system retrieves several false positives that share similar clothing styles with the query image, while our method discerns subtle differences and successfully identifies the correct individuals across various images.

This case study not only validates the robustness of our model in handling challenging ReID scenarios but also illustrates the potential of integrating LLM-derived features into traditional image processing tasks. By leveraging the advanced capabilities of LLMs for feature extraction and the power of targeted instruction-based learning, our method provides a promising direction for future ReID systems and applications in surveillance, security, and beyond.
\section{Conclusion}

In this study, we introduce MLLMReID, a novel approach for person re-identification using Multimodal Large Language Models. We tackle the overfitting issue common in instructive learning by employing a Common Instruction strategy, simplifying instructions to preserve model diversity and improve generalization. Additionally, our SyncReID module innovatively uses latent image feature vectors from LLMs, directly optimizing the visual encoder and enhancing feature extraction for ReID tasks. This streamlined approach not only boosts the effectiveness of MLLMs in ReID but also sets a foundation for future multimodal learning advancements. 

The success of our MLLMReID framework in person re-identification paves the way for future research. Key among these is the potential application of the Common Instruction strategy to a wider array of multimodal tasks, including object detection, scene understanding, and non-visual modalities.

\bibliographystyle{named}
\bibliography{ijcai24}

\begin{thebibliography}{}

\bibitem[\protect\citeauthoryear{Bao \bgroup \em et al.\egroup }{2023}]{Bao2023LearningTP}
Li-Na Bao, Longhui Wei, Xiaoyu Qiu, Wen gang Zhou, Houqiang Li, and Qi~Tian.
\newblock Learning transferable pedestrian representation from multimodal information supervision.
\newblock {\em ArXiv}, abs/2304.05554, 2023.

\bibitem[\protect\citeauthoryear{Brohan \bgroup \em et al.\egroup }{2023}]{Brohan2023RT2VM}
Anthony Brohan, Noah Brown, Justice Carbajal, Yevgen Chebotar, Krzysztof Choromanski, Tianli Ding, Danny Driess, Chelsea Finn, Peter~R. Florence, Chuyuan Fu, Montse~Gonzalez Arenas, Keerthana Gopalakrishnan, Kehang Han, Karol Hausman, Alexander Herzog, Jasmine Hsu, Brian Ichter, Alex Irpan, Nikhil~J. Joshi, Ryan~C. Julian, Dmitry Kalashnikov, Yuheng Kuang, Isabel Leal, Sergey Levine, Henryk Michalewski, Igor Mordatch, Karl Pertsch, Kanishka Rao, Krista Reymann, Michael~S. Ryoo, Grecia Salazar, Pannag~R. Sanketi, Pierre Sermanet, Jaspiar Singh, Anika Singh, Radu Soricut, Huong Tran, Vincent Vanhoucke, Quan~Ho Vuong, Ayzaan Wahid, Stefan Welker, Paul Wohlhart, Ted Xiao, Tianhe Yu, and Brianna Zitkovich.
\newblock Rt-2: Vision-language-action models transfer web knowledge to robotic control.
\newblock {\em ArXiv}, abs/2307.15818, 2023.

\bibitem[\protect\citeauthoryear{Chen \bgroup \em et al.\egroup }{2018}]{Chen2018ImprovingDV}
Dapeng Chen, Hongsheng Li, Xihui Liu, Yantao Shen, Zejian Yuan, and Xiaogang Wang.
\newblock Improving deep visual representation for person re-identification by global and local image-language association.
\newblock {\em ArXiv}, abs/1808.01571, 2018.

\bibitem[\protect\citeauthoryear{Chen \bgroup \em et al.\egroup }{2019}]{chen2019abd}
Tianlong Chen, Shaojin Ding, Jingyi Xie, Ye~Yuan, Wuyang Chen, Yang Yang, Zhou Ren, and Zhangyang Wang.
\newblock Abd-net: Attentive but diverse person re-identification.
\newblock In {\em Proceedings of the IEEE/CVF International Conference on Computer Vision}, pages 8351--8361, 2019.

\bibitem[\protect\citeauthoryear{Chen \bgroup \em et al.\egroup }{2020}]{chen2020salience}
Xuesong Chen, Canmiao Fu, Yong Zhao, Feng Zheng, Jingkuan Song, Rongrong Ji, and Yi~Yang.
\newblock Salience-guided cascaded suppression network for person re-identification.
\newblock In {\em Proceedings of the IEEE/CVF Conference on Computer Vision and Pattern Recognition}, pages 3300--3310, 2020.

\bibitem[\protect\citeauthoryear{Chen \bgroup \em et al.\egroup }{2023}]{Chen2023TowardsMP}
Cuiqun Chen, Mang Ye, and Ding Jiang.
\newblock Towards modality-agnostic person re-identification with descriptive query.
\newblock {\em 2023 IEEE/CVF Conference on Computer Vision and Pattern Recognition (CVPR)}, pages 15128--15137, 2023.

\bibitem[\protect\citeauthoryear{Ci \bgroup \em et al.\egroup }{2023}]{Ci2023UniHCPAU}
Yuanzheng Ci, Yizhou Wang, Meilin Chen, Shixiang Tang, Lei Bai, Feng Zhu, Rui Zhao, Fengwei Yu, Donglian Qi, and Wanli Ouyang.
\newblock Unihcp: A unified model for human-centric perceptions.
\newblock {\em ArXiv}, abs/2303.02936, 2023.

\bibitem[\protect\citeauthoryear{Ding \bgroup \em et al.\egroup }{2021}]{Ding2021SemanticallySN}
Zefeng Ding, Changxing Ding, Zhiyin Shao, and Dacheng Tao.
\newblock Semantically self-aligned network for text-to-image part-aware person re-identification.
\newblock {\em ArXiv}, abs/2107.12666, 2021.

\bibitem[\protect\citeauthoryear{Dosovitskiy \bgroup \em et al.\egroup }{2020}]{Dosovitskiy2020AnII}
Alexey Dosovitskiy, Lucas Beyer, Alexander Kolesnikov, Dirk Weissenborn, Xiaohua Zhai, Thomas Unterthiner, Mostafa Dehghani, Matthias Minderer, Georg Heigold, Sylvain Gelly, Jakob Uszkoreit, and Neil Houlsby.
\newblock An image is worth 16x16 words: Transformers for image recognition at scale.
\newblock {\em ArXiv}, abs/2010.11929, 2020.

\bibitem[\protect\citeauthoryear{Dosovitskiy \bgroup \em et al.\egroup }{2021}]{DBLP:conf/iclr/DosovitskiyB0WZ21}
Alexey Dosovitskiy, Lucas Beyer, Alexander Kolesnikov, Dirk Weissenborn, Xiaohua Zhai, Thomas Unterthiner, Mostafa Dehghani, Matthias Minderer, Georg Heigold, Sylvain Gelly, Jakob Uszkoreit, and Neil Houlsby.
\newblock An image is worth 16x16 words: Transformers for image recognition at scale.
\newblock In {\em 9th International Conference on Learning Representations, {ICLR} 2021, Virtual Event, Austria, May 3-7, 2021}. OpenReview.net, 2021.

\bibitem[\protect\citeauthoryear{He \bgroup \em et al.\egroup }{2021}]{he2021transreid}
Shuting He, Hao Luo, Pichao Wang, Fan Wang, Hao Li, and Wei Jiang.
\newblock Transreid: Transformer-based object re-identification.
\newblock In {\em Proceedings of the IEEE/CVF International Conference on Computer ViCVPR'21sion}, pages 15013--15022, 2021.

\bibitem[\protect\citeauthoryear{He \bgroup \em et al.\egroup }{2023}]{He2023RegionGA}
Shuting He, Weihua Chen, Kai Wang, Haowen Luo, F.~Wang, Wei Jiang, and Henghui Ding.
\newblock Region generation and assessment network for occluded person re-identification.
\newblock {\em IEEE Transactions on Information Forensics and Security}, 19:120--132, 2023.

\bibitem[\protect\citeauthoryear{Hinton \bgroup \em et al.\egroup }{2015}]{Hinton2015DistillingTK}
Geoffrey~E. Hinton, Oriol Vinyals, and Jeffrey Dean.
\newblock Distilling the knowledge in a neural network.
\newblock {\em ArXiv}, abs/1503.02531, 2015.

\bibitem[\protect\citeauthoryear{Hu \bgroup \em et al.\egroup }{2021}]{Hu2021LoRALA}
J.~Edward Hu, Yelong Shen, Phillip Wallis, Zeyuan Allen-Zhu, Yuanzhi Li, Shean Wang, and Weizhu Chen.
\newblock Lora: Low-rank adaptation of large language models.
\newblock {\em ArXiv}, abs/2106.09685, 2021.

\bibitem[\protect\citeauthoryear{Li \bgroup \em et al.\egroup }{2014}]{li2014deepreid}
Wei Li, Rui Zhao, Tong Xiao, and Xiaogang Wang.
\newblock Deepreid: Deep filter pairing neural network for person re-identification.
\newblock In {\em Proceedings of the IEEE conference on computer vision and pattern recognition}, pages 152--159, 2014.

\bibitem[\protect\citeauthoryear{Li \bgroup \em et al.\egroup }{2017}]{Li2017PersonSW}
Shuang Li, Tong Xiao, Hongsheng Li, Bolei Zhou, Dayu Yue, and Xiaogang Wang.
\newblock Person search with natural language description.
\newblock {\em 2017 IEEE Conference on Computer Vision and Pattern Recognition (CVPR)}, pages 5187--5196, 2017.

\bibitem[\protect\citeauthoryear{Li \bgroup \em et al.\egroup }{2023}]{Li2023MultimodalFM}
Chunyuan Li, Zhe Gan, Zhengyuan Yang, Jianwei Yang, Linjie Li, Lijuan Wang, and Jianfeng Gao.
\newblock Multimodal foundation models: From specialists to general-purpose assistants.
\newblock {\em ArXiv}, abs/2309.10020, 2023.

\bibitem[\protect\citeauthoryear{Liao and Shao}{2020}]{liao2020interpretable}
Shengcai Liao and Ling Shao.
\newblock Interpretable and generalizable person re-identification with query-adaptive convolution and temporal lifting.
\newblock In {\em European Conference on Computer Vision}, pages 456--474. Springer, 2020.

\bibitem[\protect\citeauthoryear{Liao and Shao}{2021}]{Liao2021TransMatcherDI}
Shengcai Liao and Ling Shao.
\newblock Transmatcher: Deep image matching through transformers for generalizable person re-identification.
\newblock In {\em Neural Information Processing Systems}, 2021.

\bibitem[\protect\citeauthoryear{Liao and Shao}{2022}]{Liao_2022_CVPR}
Shengcai Liao and Ling Shao.
\newblock Graph sampling based deep metric learning for generalizable person re-identification.
\newblock In {\em Proceedings of the IEEE/CVF Conference on Computer Vision and Pattern Recognition (CVPR)}, pages 7359--7368, June 2022.

\bibitem[\protect\citeauthoryear{Liu \bgroup \em et al.\egroup }{2023}]{Liu2023VisualIT}
Haotian Liu, Chunyuan Li, Qingyang Wu, and Yong~Jae Lee.
\newblock Visual instruction tuning.
\newblock {\em ArXiv}, abs/2304.08485, 2023.

\bibitem[\protect\citeauthoryear{Miao \bgroup \em et al.\egroup }{2019}]{miao2019pose}
Jiaxu Miao, Yu~Wu, Ping Liu, Yuhang Ding, and Yi~Yang.
\newblock Pose-guided feature alignment for occluded person re-identification.
\newblock In {\em Proceedings of the IEEE/CVF international conference on computer vision}, pages 542--551, 2019.

\bibitem[\protect\citeauthoryear{Qi \bgroup \em et al.\egroup }{2023}]{Qi2023AnID}
Baoguang Qi, Yi~Chen, Qiang Liu, Xiaohai He, Linbo Qing, Raymond~Edward Sheriff, and Honggang Chen.
\newblock An image–text dual-channel union network for person re-identification.
\newblock {\em IEEE Transactions on Instrumentation and Measurement}, 72:1--16, 2023.

\bibitem[\protect\citeauthoryear{Ristani \bgroup \em et al.\egroup }{2016}]{ristani2016performance}
Ergys Ristani, Francesco Solera, Roger Zou, Rita Cucchiara, and Carlo Tomasi.
\newblock Performance measures and a data set for multi-target, multi-camera tracking.
\newblock In {\em European conference on computer vision}, pages 17--35. Springer, 2016.

\bibitem[\protect\citeauthoryear{Shu \bgroup \em et al.\egroup }{2022}]{Shu2022SeeFS}
Xiujun Shu, Wei Wen, Haoqian Wu, Keyun Chen, Yi-Zhe Song, Ruizhi Qiao, Bohan Ren, and Xiao Wang.
\newblock See finer, see more: Implicit modality alignment for text-based person retrieval.
\newblock In {\em ECCV Workshops}, 2022.

\bibitem[\protect\citeauthoryear{Sun \bgroup \em et al.\egroup }{2018}]{sun2018beyond}
Yifan Sun, Liang Zheng, Yi~Yang, Qi~Tian, and Shengjin Wang.
\newblock Beyond part models: Person retrieval with refined part pooling (and a strong convolutional baseline).
\newblock In {\em Proceedings of the European conference on computer vision (ECCV)}, pages 480--496, 2018.

\bibitem[\protect\citeauthoryear{Touvron \bgroup \em et al.\egroup }{2023}]{touvron2023llama}
Hugo Touvron, Louis Martin, Kevin Stone, Peter Albert, Amjad Almahairi, Yasmine Babaei, Nikolay Bashlykov, Soumya Batra, Prajjwal Bhargava, Shruti Bhosale, et~al.
\newblock Llama 2: Open foundation and fine-tuned chat models.
\newblock {\em arXiv preprint arXiv:2307.09288}, 2023.

\bibitem[\protect\citeauthoryear{Wang \bgroup \em et al.\egroup }{2018}]{wang2018learning}
Guanshuo Wang, Yufeng Yuan, Xiong Chen, Jiwei Li, and Xi~Zhou.
\newblock Learning discriminative features with multiple granularities for person re-identification.
\newblock In {\em Proceedings of the 26th ACM international conference on Multimedia}, pages 274--282, 2018.

\bibitem[\protect\citeauthoryear{Wang \bgroup \em et al.\egroup }{2020}]{wang2020high}
Guan'an Wang, Shuo Yang, Huanyu Liu, Zhicheng Wang, Yang Yang, Shuliang Wang, Gang Yu, Erjin Zhou, and Jian Sun.
\newblock High-order information matters: Learning relation and topology for occluded person re-identification.
\newblock In {\em Proceedings of the IEEE/CVF Conference on Computer Vision and Pattern Recognition}, pages 6449--6458, 2020.

\bibitem[\protect\citeauthoryear{Wei \bgroup \em et al.\egroup }{2018}]{wei2018person}
Longhui Wei, Shiliang Zhang, Wen Gao, and Qi~Tian.
\newblock Person transfer gan to bridge domain gap for person re-identification.
\newblock In {\em Proceedings of the IEEE conference on computer vision and pattern recognition}, pages 79--88, 2018.

\bibitem[\protect\citeauthoryear{Xiang \bgroup \em et al.\egroup }{2023}]{Xiang2023LearningRV}
Suncheng Xiang, Jingsheng Gao, Mengyuan Guan, Jiacheng Ruan, Chengfeng Zhou, Ting Liu, Dahong Qian, and Yuzhuo Fu.
\newblock Learning robust visual-semantic embedding for generalizable person re-identification.
\newblock {\em ArXiv}, abs/2304.09498, 2023.

\bibitem[\protect\citeauthoryear{Ye \bgroup \em et al.\egroup }{2020}]{Ye2020DeepLF}
Mang Ye, Jianbing Shen, Gaojie Lin, Tao Xiang, Ling Shao, and Steven C.~H. Hoi.
\newblock Deep learning for person re-identification: A survey and outlook.
\newblock {\em IEEE Transactions on Pattern Analysis and Machine Intelligence}, 44:2872--2893, 2020.

\bibitem[\protect\citeauthoryear{Yin \bgroup \em et al.\egroup }{2023}]{Yin2023ASO}
Shukang Yin, Chaoyou Fu, Sirui Zhao, Ke~Li, Xing Sun, Tong Xu, and Enhong Chen.
\newblock A survey on multimodal large language models.
\newblock {\em ArXiv}, abs/2306.13549, 2023.

\bibitem[\protect\citeauthoryear{Zhao \bgroup \em et al.\egroup }{2021}]{zhao2021learning}
Yuyang Zhao, Zhun Zhong, Fengxiang Yang, Zhiming Luo, Yaojin Lin, Shaozi Li, and Nicu Sebe.
\newblock Learning to generalize unseen domains via memory-based multi-source meta-learning for person re-identification.
\newblock In {\em Proceedings of the IEEE/CVF Conference on Computer Vision and Pattern Recognition}, pages 6277--6286, 2021.

\bibitem[\protect\citeauthoryear{Zheng \bgroup \em et al.\egroup }{2015}]{zheng2015scalable}
Liang Zheng, Liyue Shen, Lu~Tian, Shengjin Wang, Jingdong Wang, and Qi~Tian.
\newblock Scalable person re-identification: A benchmark.
\newblock In {\em Proceedings of the IEEE international conference on computer vision}, pages 1116--1124, 2015.

\bibitem[\protect\citeauthoryear{Zhou \bgroup \em et al.\egroup }{2019}]{zhou2019omni}
Kaiyang Zhou, Yongxin Yang, Andrea Cavallaro, and Tao Xiang.
\newblock Omni-scale feature learning for person re-identification.
\newblock In {\em Proceedings of the IEEE/CVF International Conference on Computer Vision}, pages 3702--3712, 2019.

\bibitem[\protect\citeauthoryear{Zhu \bgroup \em et al.\egroup }{2020}]{zhu2020identity}
Kuan Zhu, Haiyun Guo, Zhiwei Liu, Ming Tang, and Jinqiao Wang.
\newblock Identity-guided human semantic parsing for person re-identification.
\newblock In {\em European Conference on Computer Vision}, pages 346--363. Springer, 2020.

\bibitem[\protect\citeauthoryear{Zhu \bgroup \em et al.\egroup }{2021}]{Zhu2021DSSLDS}
Aichun Zhu, Zijie Wang, Yifeng Li, Xili Wan, Jing Jin, Tian Wang, Fangqiang Hu, and Gang Hua.
\newblock Dssl: Deep surroundings-person separation learning for text-based person retrieval.
\newblock {\em Proceedings of the 29th ACM International Conference on Multimedia}, 2021.

\bibitem[\protect\citeauthoryear{Zhu \bgroup \em et al.\egroup }{2023}]{Zhu2023MiniGPT4EV}
Deyao Zhu, Jun Chen, Xiaoqian Shen, Xiang Li, and Mohamed Elhoseiny.
\newblock Minigpt-4: Enhancing vision-language understanding with advanced large language models.
\newblock {\em ArXiv}, abs/2304.10592, 2023.

\bibitem[\protect\citeauthoryear{Zhuang \bgroup \em et al.\egroup }{2020}]{zhuang2020rethinking}
Zijie Zhuang, Longhui Wei, Lingxi Xie, Tianyu Zhang, Hengheng Zhang, Haozhe Wu, Haizhou Ai, and Qi~Tian.
\newblock Rethinking the distribution gap of person re-identification with camera-based batch normalization.
\newblock In {\em European Conference on Computer Vision}, pages 140--157. Springer, 2020.

\end{thebibliography}

\end{document}